\pdfoutput=1

\documentclass[11pt]{article}

\usepackage[final]{emnlp2021}

\usepackage{times}
\usepackage{latexsym}

\usepackage[T1]{fontenc}

\usepackage[utf8]{inputenc}

\usepackage{microtype}
\usepackage{adjustbox}
\usepackage{multirow}
\usepackage{diagbox}
\usepackage{amsmath}
\usepackage{pifont}
\usepackage{graphicx}
\usepackage{amssymb}
\usepackage{arydshln}

\usepackage{booktabs}
\usepackage{caption}
\usepackage{subcaption}
\usepackage{color}
\usepackage{makecell}
\usepackage{xcolor}
\usepackage{algorithm}
\usepackage[noend]{algpseudocode}
\usepackage{pifont}
\usepackage{amsmath}
\usepackage{pdfpages}

\newcommand{\sep}{\textsc{<sep>}}

\title{Netmarble AI Center's WMT21 Automatic Post-Editing Shared Task Submission}

\makeatletter
\newcommand{\printfnsymbol}[1]{%
  \textsuperscript{\@fnsymbol{#1}}%
}
\makeatother
\author{Shinhyeok Oh$\thanks{\ \ These authors equally contributed to this work.}$\;\,, Sion Jang$\printfnsymbol{1}$, Hu Xu$\printfnsymbol{1}$, Shounan An, Insoo Oh\\Netmarble AI Center}

\begin{document}
\maketitle
\begin{abstract}
This paper describes Netmarble's submission to WMT21 Automatic Post-Editing (APE) Shared Task for the English-German language pair. First, we propose a Curriculum Training Strategy in training stages. Facebook Fair's WMT19 news translation model was chosen to engage the large and powerful pre-trained neural networks. Then, we post-train the translation model with different levels of data at each training stages. As the training stages go on, we make the system learn to solve multiple tasks by adding extra information at different training stages gradually. We also show a way to utilize the additional data in large volume for APE tasks. For further improvement, we apply Multi-Task Learning Strategy with the Dynamic Weight Average during the fine-tuning stage. To fine-tune the APE corpus with limited data, we add some related subtasks to learn a unified representation. Finally, for better performance, we leverage external translations as augmented machine translation (MT) during the post-training and fine-tuning. As experimental results show, our APE system significantly improves the translations of provided MT results by -2.848 and +3.74 on the development dataset in terms of TER and BLEU, respectively. It also demonstrates its effectiveness on the test dataset with higher quality than the development dataset. 
\end{abstract}
\section{Introduction}

Automatic Post-Editing (APE) aims to improve the quality of an existing Machine Translation (MT) system by learning from human-edited samples \cite{chatterjee-etal-2019-findings, chatterjee-etal-2020-findings}. 
With the continuous performance improvements of Neural Machine Translation (NMT) systems along with deep learning advancements, developing APE systems has faced a big challenge. Simple translation errors are hard to find in machine translation outputs, and the remaining errors are still hard to solve.
In recent years, transfer learning and data augmentation techniques have shown their efficiency when training models on datasets with limited size \cite{devlin-etal-2019-bert}. Therefore, such approaches are also adopted in APE tasks \cite{lopes-etal-2019-unbabels}.

Participants in WMT21 APE shared tasks are required to develop systems to automatically post-edit the translation outputs from an unknown MT system.
In this year, the same data has been re-post-edited to improve the quality. 
As a result of performing statistics on the development set, the evaluation scores are 19.057 and 68.79 in terms of TER and BLEU, which are much higher than the scores of last year, 31.374 and 50.37, respectively. The central distribution of TER has shifted to the left compared to last year. We find that the section in the range of 5 to 10 has the most examples, which indicates that over-correction problems should be considered during the APE tasks.
In addition, the dataset has been changed in terms of the domain (from IT to Wikipedia), which results in the change in data distribution. Therefore, directly using previous datasets or officially provided synthetic corpus~\cite{junczys-dowmunt-grundkiewicz-2016-log, negri-etal-2018-escape} to enlarge the training set of APE tasks might not be appropriate under such circumstances.
In work by \citet{yang-EtAl:2020:WMT}, considering the change of data distribution, they select to use additional MT candidates as the data augmentation method to improve feature diversity in their APE systems, which significantly improves the APE performance.

Inspired by this idea, we decided to solve the APE task as NMT alike task and utilize the external MT at the fine-tuning stage.
However, because of the limited size of the APE corpus and the improvement of MT quality, fine-tune the model only on the APE data, easily reach the performance ceiling in spite of using external translation.
To solve the aforementioned issues, existing works for other Natural Language Processing (NLP) tasks have adopted several Multi-task Learning (MTL) methods with the auxiliary task~\cite{Whang_Lee_Oh_Lee_Han_Lee_Lee_2021, oh-etal-2021-deep}.
We wondered whether it is possible to apply MTL mechanism with APE task to the fine-tuning stage since MTL trains the model to encourage representation sharing and improve generalization performance. 
Furthermore it aims to alleviate the data sparsity problem with a limited number of data in each task \cite{9392366}.
Therefore, we add some related NLP tasks along with the APE task. 
Our experiment results demonstrate that such approaches can further improve performance.

As mentioned above, large-volume data, such as news translation data and artificial synthetic data, can not be used to enlarge the APE corpus directly during the fine-tuning because of the large gap in data distribution. 
We wondered if there is a way to apply any learning method to the post-training so that we can utilize more data to train a more robust and powerful model.
In work by \cite{xu-etal-2020-curriculum}, they applied Curriculum Learning according to the difficulty of each example on a single training stage. Inspired by the research, we try to apply Curriculum Learning across multiple training stages. As the training stage increases, we make the system learn to solve the different tasks by gradually providing extra information, described in Section 3 in detail. 
Extensive experiments show the effectiveness of applying the Curriculum Learning Strategy during the training phase. 
Finally, We combined these two approaches to make our final APE system, which significantly improves the performance of the APE task.

Our APE system is built based on Transformer~\cite{NIPS2017_3f5ee243} and is post-trained on WMT21 News-Translation Data~\cite{koehn2005epc, tiedemann-2012-parallel, rozis-skadins-2017-tilde, Bhatia16, tiedemann-2012-parallel} and artificial synthetic data~\cite{junczys-dowmunt-grundkiewicz-2016-log, negri-etal-2018-escape} provided by APE Task with Curriculum Learning Strategy. 
For fine-tuning, MTL is applied with related NLP subtasks such as Part-Of-Speech (POS), Named Entity Recognition (NER), Masked Language Model (MLM), and Keep/Translate are added to the model to reduce the over-fitting as well as achieve better performance, described in Section 4 in detail. 
For better training efficiency, the Dynamic Weight Average (DWA) mechanism \cite{Liu_2019_CVPR} is applied during the MTL to keep the correct balance between these subtasks.
Here we summarize our contributions as follows:
\begin{itemize}
    \item We design Multi-task Learning Strategy (MLS) with DWA to the fine-tuning stage, which improves the training efficiency and the performance significantly.
    \item We adapt Curriculum Training Strategy (CTS) to our APE system during the post-training across the multiple training stages, which shows the effectiveness in performance. In addition, we showed a way to utilize the additional data in large volumes in APE tasks. 
\end{itemize}
\section{Base System}
\begin{figure*}[t]\centering
\includegraphics[width=\textwidth]{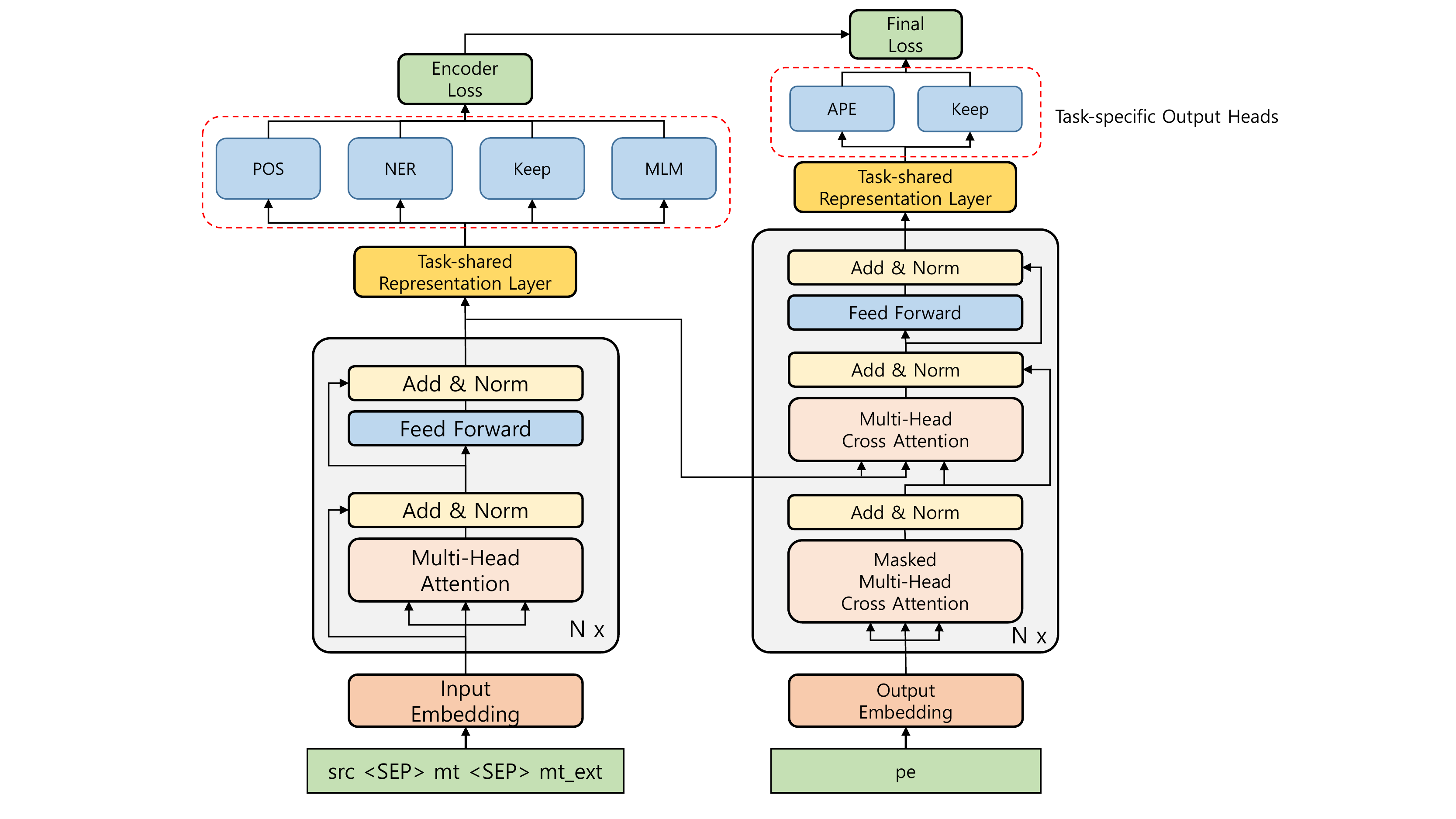}
\caption{Overall architecture}
\label{fig:a}
\vspace{-0.2cm}
\end{figure*}
\label{sec:2}
Our system is based on Facebook FAIR's WMT19 News Translation Model~\cite{ng-EtAl:2019:WMT}, which used the big Transformer~\cite{NIPS2017_3f5ee243} and provided the pre-trained weights. We use both of them as our base system. In addition, we utilize data augmentation with external MT, which has been proposed by \citet{yang-EtAl:2020:WMT} to generate the external translated sentence ($mt\_ext$) and help generate the post-editing sentence ($pe$). An input sentence $X$ that contains a source sentence ($src$), a translated sentence by the machine translation system ($mt$), and an external translated sentence ($mt\_ext$) is defined as,
\begin{equation}
\label{eq:1}
X = [src\,\sep\,mt\,\sep\,mt\_{ext}],
\end{equation}
and output a sequence, \(H = [h_{src_0}, h_{src_1}, ..., \) \\ \(h_{src_n}, h_{\sep}, h_{mt_0}, ..., h_{mt_m}, h_{\sep}, h_{mt\_{ext}_0}, ...,\\ h_{mt\_{ext}_l}]\) \(
\in \mathbb{R}^{d_h\times (n+m+l+2)}\), where \(d_h\) represents a dimension of the encoder, and \(n\), \(m\), \(l\) represents the number of tokens for \(src\), \(mt\), \(mt\_ext\), respectively. We represent the parameters of the encoder as \(\Theta_{s}\). Then, \(H\) is fed into the decoder, and the decoder target is defined as \(Y = [ pe ]\).
\section{Curriculum Training Strategy (CTS)}
CTS has been inspired by Curriculum Learning \cite{xu-etal-2020-curriculum} that is applied according to the difficulty of each example on a single training stage, which has already been applied to our baseline architecture by \citet{ng-EtAl:2019:WMT}. In addition, we propose CTS, which applied Curriculum Learning across multiple training stages. CTS aims at step-by-step learning. In an early stage, the system learns to solve easy problems or something that needs to know beforehand and complex problems or target tasks in the later stages.
\subsection{Step 1: Understanding for Machine Translation}
\begin{equation}
\label{eq:2}
X = [src],
\end{equation}
The APE task has to understand the machine translation system because the APE task modifies the \(mt\) results. Therefore, we designed the first step of the curriculum with the input as Equation~\ref{eq:2} and the target as \(pe\).

\subsection{Step 2: Learning about Post-Editing}
\begin{equation}
\label{eq:3}
X = [src\,\sep\,mt],
\end{equation}
After the first step, our system understands as the machine translation system. 
In this step, we make our system learn how to edit \(mt\) to \(pe\) with the input as Equation~\ref{eq:3} and the target as \(pe\).
\subsection{Step 3: Post-Editing with External MT}
For the second step, our system learns about the post-editing mechanism. 
In this step, we make the system learn to take the External MT into account with the input as Equation~\ref{eq:1} and the target as \(pe\).
\subsection{Fine-Tuning}
Finally, we fine-tune the APE system using the data given in the challenge with the input as Equation~\ref{eq:1} and the target as \(pe\).
\section{Multi-task Learning Strategy (MLS)}
Existing works for MTL propose jointly learning methods among related tasks. 
MTL aims to improve the generalization performance of the whole tasks by sharing knowledge representations of other tasks and can also alleviate the data sparsity problem where each task has limited labeled data~\cite{9392366}. 
Therefore, we utilize MLS for our system because WMT21 APE shared task provides only 7,000 train sentences.
In NMT, existing works for MTL applied POS, NER, or MLM as subtasks and provided improved results~\cite{chatterjee-EtAl:2017:WMT1, wang-etal-2020-multi}. 
Despite the impressive results, they applied only a few subtasks, such as one or two.
Since we defined the APE task as NMT alike problem in our work, it would be helpful to leverage these subtasks into our work to achieve better performance.
We find out that all these subtasks are cooperative with each other and benefit our system. 
Inspired by the word-level quality estimation task, we also add the Keep/Translate classification tasks for encoder and decoder to handle the high-quality APE task, which is described in Section~\ref{MLS:KT} in detail.
Since utilizing multiple subtasks, we have to consider the loss ratio between these subtasks. 
In our work, we apply the Dynamic Weight Average method described in \citet{Liu_2019_CVPR}, and more details are described in Section \ref{MLS:DWA}.
Our final system based on the model post-trained using CTS with fine-tuning the APE data with MLS. 

\begin{figure*}[t]\centering
\includegraphics[width=\textwidth]{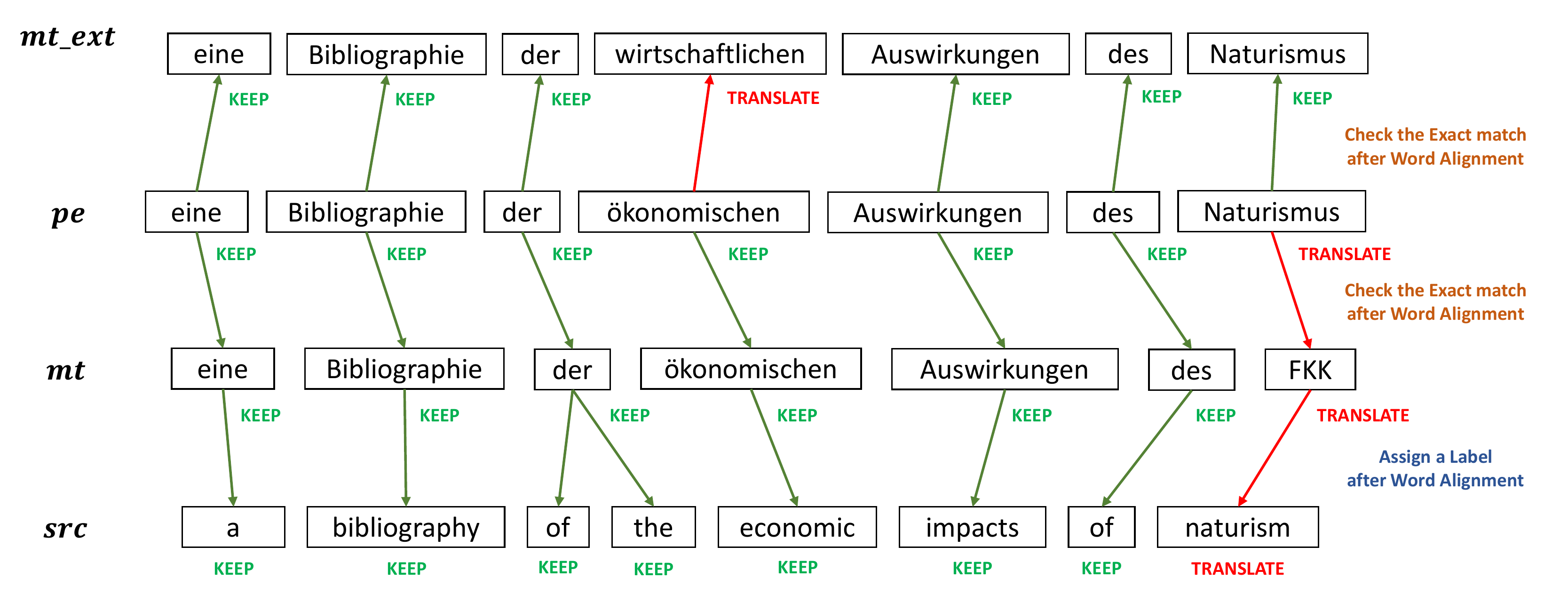}
\caption{A label generation example in the Keep/Translate task}
\label{fig:b}
\vspace{-0.2cm}
\end{figure*}

\subsection{Architecture}
Our architecture is described in Figure~\ref{fig:a}. The overall flow of the APE task is the same in Section~\ref{sec:2}. In this section, we explain five auxiliary subtasks consisting of POS, NER, MLM, Keep/Translate for the encoder, and Keep/Translate for the decoder. For the encoder, the encoding vector \(H\) is fed into Task-shared Representation Layer in Figure~\ref{fig:a} like a Fully-connected Neural Network (FNN), and the output is represented as,
\begin{equation}
\label{eq:4}
\begin{split}
H_{s} &= (W_{1}H + b_{1}), 
\end{split}
\end{equation}
where \(W_{1} \in \mathbb{R}^{d^{s}_{h}\times d_h}\), and \(d^{s}_{h}\) represents a dimension of the Task-shared Representation Layer.

\subsection{Subtasks}
\paragraph{POS \& NER} POS and NER task aims to predict parts of speech and named entities about an input sequence, respectively. Task-shared Representation Layer \(H_s\) is fed into Task-specific Output Heads on Figure~\ref{fig:a} like a FNN, and the output is represented as,
\begin{equation}
\label{eq:5}
\begin{split}
\hat{Y}^{pos} &= \text{softmax}(W_{2}H_{s} + b_{2}), 
\end{split}
\end{equation}
where \(W_{2} \in \mathbb{R}^{C_{pos}\times d_h}\) is trainable parameters and \(C_{pos}\) is the number of class of POS task. The parameters of Task-specific Output Heads for POS task are represented as \(\Theta_{pos}\). Likewise, \(\hat{Y}^{ner}\) is obtained as in Equation~\ref{eq:5} for NER task, where the parameters are represented as \(\Theta_{ner}\).

\paragraph{MLM} 
In MLM task, we copy the input tokens from \(X\) to \(X^{mlm}\), which is represented by \(X^{mlm}=\{x_1, ..., x_{n+m+l+2}\}\), where \(n\), \(m\), \(l\) represents the number of tokens for \(src\), \(mt\), \(mt\_ext\), respectively. Then, we randomly mask 15\% of the tokens \(X^{mlm}\) using the special token \(mask\), and define the target as original input tokens. \(X^{mlm}\) is fed into the encoder. Then, the output representation is used to the input for Task-specific Output Heads for MLM task as,
\begin{equation}
\label{eq:6}
\begin{split}
\hat{Y}^{3} &= \text{softmax}(W_{3}H_{s} + b_{3}),\\ 
\hat{Y}^{mlm} &= \{\hat{Y}^{3}_{r} | x_{r} = mask,
\\ &\qquad \forall\ r \in \{0, ..., n+m+l+2\}\}
\end{split}
\end{equation}
where \(W_{3} \in \mathbb{R}^{C_{mlm} \times d_h}\) represents trainable parameters and \(C_{mlm}\) is the number of vocab for the encoder. The parameters of a linear projection layer are represented as \(\Theta_{mlm}\) for MLM task.

\paragraph{Keep/Translate}
\label{MLS:KT}
Considering the characteristics of the APE data with relatively low TER scores, we decide to add Keep/Translate classification subtask to both Encoder and Decoder in our APE system. Keep/Translate subtask aims to predict the labels of the input sequence, where is \(\hat Y^{kt} \in \{Keep, Translate\}\). 
In this subtask, each token in the input will be labeled with \(Keep\) or \(Translate\). 
For label generation, we apply to the pair of \(src\)-\(mt\) and \(src\)-\(mt\). First, we use SimAlign \cite{jalili-sabet-etal-2020-simalign} to perform word alignment on the \(pe\)-\(mt\) pair. To each aligned word pair, we labeled them with Keep if they are equal. Otherwise, they will be marked as Translate. 
As for the pair of \(src\)-\(mt\), we also do word alignment to find the correspondence between the source and target side. On the \(src\) side, the tokens are labeled with the same name as the corresponding words on the \(mt\) side. 
In our case, the same procedure on \(pe\)-\(mt\) is conducted for the pair of \(mt\_ext\) and \(pe\) because we use the \(mt\_ext\) as our data augmentation method.
Figure~\ref{fig:b} shows an example of label generation in the Keep/Translate task for better understanding. 
The output is represented as,
\begin{equation}
\label{eq:55}
\begin{split}
\hat{Y}^{kt} &= \text{softmax}(W_{4}H_{s} + b_{4}), 
\end{split}
\end{equation}
where \(W_{4} \in \mathbb{R}^{C_{kt}\times d_h}\) is trainable parameters and \(C_{kt}\) is the number of class of Keep/Translate task. The parameters of Task-specific Output Heads for Keep/Translate task are represented as \(\Theta_{kt}\) and \(\hat{Y}^{kt}\) is obtained as in Equation~\ref{eq:55} for Keep/Translate task.

\subsection{Loss}
As described above, five subtasks are used in our system, and most of them have data with imbalanced labels. 
The imbalanced ratio reaches 1:2160, 1:15, and 1:6 between minority and majority classes in POS Tagger, NER, and Keep/Translate subtasks, respectively.
With such imbalanced data, the Cross-Entropy loss used in classification problems may result in performance degradation in some tasks. To improve the performance, the Focal loss \cite{8237586} is considered as an alternative candidate because a Focal Loss function addresses class imbalance during training in tasks. It applies a modulating term to the cross-entropy loss in order to focus on learning the hard negative examples. It reduces the relative loss for well-classified examples (\(p_{\text{t}} \) > 0.5), putting more focus on hard, misclassified examples. Equation~\ref{eq:focalloss} describes the Focal Loss, where \(p_{\text{t}} \) is the probability of each class predicted by the model and \(\gamma\) represents the focusing parameter. 
Considering the imbalanced property of each task, we apply the Focal Loss to three of our subtasks, such as POS Tagger, NER, and Keep/Translate in the decoder.

\begin{equation}
\label{eq:focalloss}
{FL(p_{\text{t}})} = -(1-p_{\text{t}}) ^ {\! \gamma} log(p_{\text{t}})
\end{equation}

Class Balanced Loss is designed to use a re-weighting scheme that uses the effective number of samples for each class to re-balance the loss, thereby yielding a class-balanced loss \cite{8953804}.
As the number of samples increases, there is information overlap among data. Therefore, the marginal benefit that a model can extract from the data diminishes. 
The effective number of samples, which played as the expected volume of samples, is used to capture the diminishing marginal benefits by using more data points of a class. 
For Keep/Translate task in the decoder, it just considered the PE as input, so we applied the Focal Loss to the subtask. 
However, for Keep/Translate task in the encoder, as one of the data augmentation methods, the external MT is also considered as input along with the \(src\) and \(mt\).
As the information of input increases, we think it may cause information overlap among data because \(mt\) and the \(mt\_ext\) have the most in common. Therefore, we apply the Class-Balanced Loss as our loss function in Keep/Translate subtask in the encoder.
Equation~\ref{eq:classbalancedloss} describes the Class-Balanced Loss (\(L_{cb}\)), where \(C\) is the total number of classes, \(z_{\text{y}} \) is the output from the model for class \(y\), \({n_y}\) is the number of samples in the ground-truth class and \(\beta\) \(\in\) [0, 1) is a hyperparameter which can be calculated in Equation~\ref{eq:classbalancedlossterm}. In Equation~\ref{eq:classbalancedlossterm}, \(i\) denotes the class index, \(i \in \{1,2,...,C\}\), and \(N\) is the number of samples.

As for the MLM task, since it does not suffer from the data imbalance problem, we use the Cross-Entropy loss in our work as other works do.

\begin{equation}
\label{eq:classbalancedlossterm}
\begin{alignedat}{2}
N_i &= N,\\
\beta_i &= \beta = (N-1) / N
\end{alignedat}
\end{equation}
\begin{equation}
\label{eq:classbalancedloss}
L_{cb} = -\frac{1 - \beta}{1 - \beta^{n_y}} \log\left(\frac{\exp(z_y)}{\sum_{j=1}^C \exp(z_j)}\right)
\end{equation}

\subsection{Dynamic Weight Average}
\label{MLS:DWA}
For most Multi-Task learning networks, it's difficult to find the best ratio between each task in subtasks manually. Therefore, we apply the Dynamic Weight Average (DWA) \cite{Liu_2019_CVPR} to our work, which adapts the task weighting over time by considering the rate of change of the loss for each task. 

Equations~\ref{eq:mtlWeightForTraining} and~\ref{eq:eachMtlLossByEpoch} describe DWA. Here, \(\lambda_{\text{k}}(\cdot) \) represents the weighting for task \(k\), \(w_{\text{k}}(\cdot) \) calculates the relative descending loss rate for each task in each epoch, \(t\) is an iteration index, and \(T\) represents a temperature that controls the softness of task weighting.
\(\mathcal{L} \) in Equation~\ref{eq:eachMtlLossByEpoch} is the loss value, calculated as the average loss in each epoch over several iterations.

\begin{equation}
\label{eq:mtlWeightForTraining}
\lambda_{\text{k}}{(t)} := \frac{K {exp}(\omega_\text{k}(t-1)/T)}{\sum_{i}K {exp}(\omega_\text{i}(t-1)/T)}
\end{equation}

\begin{equation}
\label{eq:eachMtlLossByEpoch}
\omega_\text{k}(t-1) = \frac{\mathcal{L}_{\text{k}}(t-1)}{\mathcal{L}_{\text{k}}(t-2)} 
\end{equation}

\subsection{Joint Learning Procedure}
All tasks are jointly trained, and the objective is defined as,
\begin{equation}
\mathcal{L} = \frac{1}{K} \sum_i^K \lambda_{\text{i}} \mathbb{L}(Y_i, f(X_i)),
\end{equation}
where \(\lambda\) is a dynamic weight determining the degree of subtasks and \(f\) is the training classifier. Note that the parameter \(K\) is the number of subtasks. \(\mathbb{L}(Y, f(X))\) is the loss of \(f\) w.r.t. the target \(Y\).
\section{Experiments}
\begin{table}[t]\centering
\begin{adjustbox}{width=0.4\textwidth}
\begin{tabular}{lcc}
\toprule
 & TER  & BLEU  \\
\midrule
CTS-best (ensemble) & \textbf{16.44} & 71.88  \\
\midrule
CTS-best (single) & 16.46 & \textbf{71.94}  \\
\ \ \ w/o\ \(step\ 2\)  & 16.70 & 71.84  \\
\ \ \ w/o\ \(step\ 3\)  & 17.33 & 70.24  \\
\ \ \ w/o\ \(step\ 2\)\ \&\ 3  & 17.28 & 70.88  \\
\midrule
\(baseline\) & 19.06 & 68.79  \\
\bottomrule
\end{tabular}

\end{adjustbox}

\caption{CTS results on WMT21 APE development dataset. CTS-best (ensemble) is built by two similar single models. we submitted CTS-best (ensemble) as CONTRASTIVE result.}
\label{table:curriculum}
\vspace{-0.2cm}
\end{table}
\subsection{Datasets}
Following existing works, we utilize additional resources~\cite{junczys-dowmunt-grundkiewicz-2016-log, negri-etal-2018-escape}, which have source sentences (\(src\)), machine translation sentences (\(mt\)), and post-editing sentences (\(pe\)). Moreover, we also utilize some of News-Translation data for the WMT21~\cite{koehn2005epc, tiedemann-2012-parallel, rozis-skadins-2017-tilde, Bhatia16, tiedemann-2012-parallel}, which has source sentences (\(src\)) and translated sentences that can be used as \(pe\). For evaluation and fine-tuning, we use the data for WMT21 automatic post-editing shared task. Moreover, we utilize translated sentences using Google Translate and Quality Estimation NMT Model~\cite{fomicheva-etal-2020-unsupervised}. The former is used to make \(mt\_ext\) from the additional resources and the data for WMT21 automatic post-editing. The latter is used to make \(mt\) from News-Translation data.
We filtered all the training data based on and number checking logic, which filters the pairs with different numbers in source and target side.

\subsection{Experimental Settings}

For the first step of CTS, we utilize WMT19 en-de weights by Fairseq~\cite{ng-EtAl:2019:WMT}. In the second step, we utilize News-Translation data with translated sentences with Quality Estimation NMT Model as \(mt\). In the third step, we make our system learn with \citet{junczys-dowmunt-grundkiewicz-2016-log, negri-etal-2018-escape} and Google Translate as \(mt\_ext\). Finally, when learning the fine-tuning step, which contains MLS, we utilize the data for WMT21 Automatic Post-Editing shared task.

\begin{table}[t]\centering
\begin{adjustbox}{width=0.35\textwidth}
    \begin{tabular}{lccc}
        \toprule
            & TER  & BLEU  \\
        \midrule
            MLS\ \ w\ DWA  & \textbf{16.21} & \textbf{72.53}  \\
            MLS\ \ w/o\ DWA  & 16.37 & 72.34  \\
        \bottomrule
    \end{tabular}

\end{adjustbox}

\caption{Ablation analysis of DWA on the WMT21 APE development dataset.}
\label{tab:2}
\vspace{-0.2cm}
\end{table}
\begin{table}[t]\centering
\begin{adjustbox}{width=0.4\textwidth}
    \begin{tabular}{lccccccc}
        \toprule
            & TER  & BLEU  \\
        \midrule
            Vanilla & 16.71 & 71.75  \\
        \midrule
            \quad\, w/ POS & 16.49 & 72.12  \\
            \quad\, w/ NER & 16.52 & 72.19  \\
            \quad\, w/ MLM & 16.55 & 72.00  \\
            \quad\, w/ Keep/Translate & 16.45 & 72.32  \\
        \midrule
            Fine-tuned\ with\ MLS  & \textbf{16.21} & \textbf{72.53}  \\
        \bottomrule
    \end{tabular}
\end{adjustbox}
    \caption{The Multi-task Learning results on WMT21 APE validation dataset. Fine-tuned with MLS using all subtasks model is submitted as PRIMARY result.}
    \label{tab:3}
\vspace{-0.2cm}
\end{table}
\subsection{Results: CTS}
To study the effectiveness of CTS, we conduct ablation experiments on WMT21 Automatic Post-Editing development dataset. We set the \(baseline\), which is a system that leaves all the test instances unmodified. As shown in Table~\ref{table:curriculum}, we can observe that the \(step\ 3\) is more effective than the \(step\ 2\), and that using only \(step\ 2\) doesn't help APE. As our system is learning step by step with CTS, it allows that our system has strengths in the APE task.
\begin{table*}[t]
\centering
\resizebox{0.65\textwidth}{!}{
    \begin{tabular}{lccc}
        \toprule
            & Avg  & Avg z  \\
        \midrule
            Netmarble\_PRIMARY & \textbf{79.82} & \textbf{0.144}  \\
        \midrule
            Netmarble\_CONTRASTIVE & 78.52 & 0.095  \\
        \midrule
            PVIE\_PRIMARY~\cite{Sharma-etal-2021-adapting} & 76.85 & 0.02 \\
            PVIE\_CONTRASTIVE~\cite{Sharma-etal-2021-adapting} & 76.67 & 0.011 \\
        \midrule
            \(baseline\)~\cite{Akhbardeh-etal-2021-finding} & 69.68 & -0.27  \\
        \bottomrule
    \end{tabular}}
\caption{Official results on WMT21 APE human evaluation~\cite{Akhbardeh-etal-2021-finding}. We utilized both CTS and MLS in Netmarble\_PRIMARY submission and only utilized CTS in Netmarble\_CONTRASTIVE submission. PVIE\_PRIMARY is the ensemble model utilizing domain and task adaptation methods, and PVIE\_CONTRASTIVE is the single model. Systems ordered by DA score; systems within a cluster are considered tied; lines indicate clusters according to Wilcoxon rank-sum test p < 0.05.}
\label{tab:5}
\vspace{-0.2cm}
\end{table*}
\subsection{Results: MLS}
Table~\ref{tab:2} presents the ablation analysis about DWA when fine-tuning with MLS on WMT21 APE development set. From the result, we can observe that MLS with DWA has better performance than the one without applying it. For that reason, we adopt DWA at a fine-tuning stage with MLS in our APE Task.

To find the best combination of subtasks in MLS, we conducted an ablation analysis on the same development dataset. Vanilla in the table is a system without adding any subtasks. We add the subtasks one by one during the fine-tuning to see the effect of each subtask on the performance. As shown in Table~\ref{tab:3}, the one using all the subtasks performs best among all the combinations, which means that these subtasks are cooperative in the APE task.

\subsection{Official Results}
Table~\ref{tab:5} shows the official results of our submissions on the test dataset. \citet{Akhbardeh-etal-2021-finding} noticed Netmarble\_PRIMARY submission which contains CTS and MLS significantly outperforms all other submissions and can be declared as the single winner of the WMT21 APE task. More details are described in their finding paper.

\subsection{Implementation Details}
We set the batch size to 256 for the \(step\ 2\) and \(step\ 3\) in CTS at each GPU, 16 for the fine-tuning and MLS. We set the initial learning rate to 1e-4 using scheduler in Fairseq~\cite{ng-EtAl:2019:WMT} for all experiments. The average runtime of one epoch for each approach was about 360 minutes for the \(step\ 2\), 90 minutes for the  \(step\ 3\), and 40 seconds for MLS. We train our models using AdamW~\cite{DBLP:conf/iclr/LoshchilovH19} optimizer and conduct experiments with 16 Tesla A100 GPUs for CTS, Tesla V100 GPU for MLS.
\section{Conclusion}
In this paper, we propose an APE system based on CTS and MLS. 
CTS allows understanding between machine translation and automatic post-editing, and shows a way using additional data in large volume in APE task.
MLS learns a shared unified representation from related subtasks to improve the performance. 
We submitted the system, which Fine-tunes with MLS, as our primary version and the ensembled CTS as our contrastive version.
The experimental results show that our system is able to effectively detect and correct the errors made by a high-quality NMT system, improving the score by -2.848 and +3.74 on the development dataset in terms of TER and BLEU, respectively. Our proposed methods also achieved performance improvement on the test dataset with higher quality.

\bibliography{emnlp2021}
\bibliographystyle{acl_natbib}

\end{document}